# Non - invasive modelling methodology for the diagnosis of Coronary Artery Disease using Fuzzy Cognitive Maps

Ioannis D. Apostolopoulos [1*], Peter P. Groumpos [2]

[1]  Department of Medical Physics, School of Medicine, University of Patras, 26504 Patras, Greece

[2]  Electrical and Computer Engineering Department, University of Patras, 26504 Patras, Greece

\*   Correspondence: ece7216@upnet.gr

**Abstract:** Cardiovascular Diseases (CVD) and strokes produce immense health and economic burdens globally. Coronary Artery Disease (CAD) is the most common type of cardiovascular disease. Coronary Angiography, which is an invasive treatment, is also the standard procedure for diagnosing CAD. In this work, we illustrate a Medical Decision Support System for the prediction of Coronary Artery Disease (CAD) utilizing Fuzzy Cognitive Maps (FCMs). FCMs are a promising modeling methodology, based on human knowledge, capable of dealing with ambiguity and uncertainty, and learning how to adapt to the unknown or changing environment. The newly proposed MDSS is developed using the basic notions of Fuzzy Logic and Fuzzy Cognitive Maps, with some adjustments to improve the results. The proposed model, tested on a labelled CAD dataset of 303 patients, obtains an accuracy of 78.2% outmatching several state-of-the-art classification algorithms.

# 1 Introduction

The Coronary Artery Disease (CAD) is one of the most common causes of death worldwide. CAD develops when the major blood vessels (coronary arteries) that supply the heart with blood, oxygen, and nutrients become damaged or diseased. The non-invasive and accurate diagnosis of CAD is a challenging task. The standard diagnostic tests and factors affecting the risk of CAD do not guarantee a reliable diagnosis. The enormous number of factors contributing to CAD and the complex connections between them makes it difficult for the doctors to handle the clinical data and the conditions. Hence, a Medical Decision Support System (MDSS) could provide a second opinion on the matter and possibly improve the significance of the final medical report.

Developing Decision Support Systems is one of the most notable and vital efforts, and their use helps in solving daily problems in different areas. Examples of applying them are found in the health, security and telecommunications sectors (Jaspers et al. 2011).

Diagnosing the Coronary Artery Disease (CAD) in a non – invasive way is not a new challenge (Sintchenko et al. 2007). Several works and proposals are demonstrated over the years to achieve high accuracy (Acharya et al. 2017). The majority of the proposals imply data mining and machine learning techniques to a variety of datasets. Pattern recognition methods are also utilized in order to obtain information from medical images such as ECG or Myocardial Perfusion Imaging (also known as Scintigraphy). We illustrate few characteristic recent works. A summary of every technique and dataset that was employed can be found in (Alizadehsani et al. 2019).

Most of recent research works usilize large number of patients. A hybrid system of rough set and neural network has been proposed by Liping and Lingyun (Liping and Lingyun 2005). A research work proposed by Ordonez (2006) that uses association rules instead of decision rules for heart disease prediction is proposed in (Ordonez 2006). Rajkumar and Reena (Rajkumar and Reena 2010) used decision tree and Naïve Bayes algorithms on the public University of California Irvine (UCI) dataset and reached 52.33%. In their work, El-Bialy (El-Bialy 2015) applied an integration of the results of the machine learning analysis applied on different data sets targeting the CAD disease. Other works include techniques such as neural networks (Patil 2009), the

Bayesian model and decision tree, support vector machine (Lapuerta 1995), and the naive Bayes classifier (Kampouraki 2009).

Babagolu et al. (2010) employed data from Treadmill Exercise Tests and a Support Vector Machine (SVM) for the classification task, obtaining 81.46% accuracy. Different feature selection methods such as filter method (Setiawan 2009), genetic algorithm (Arabasadi 2017), and numerical and nominal attribute selection have been used for artery stenosis disease prediction. For the diagnosis of CAD based on evidence, Setiawan et al. (Setiawan 2009) have developed a fuzzy decision support system. The CAD data sets obtained from the University of California Irvine (UCI) are utilized.

This work aims to enhance and to further develop an already proposed Medical Decision Support System for the diagnosis of Coronary Artery Disease. The first proposal (Apostolopoulos et al. 2017) was published in 2017. In their work, the authors utilized Fuzzy Cognitive Maps to mimic the medical experts' way of thinking, obtaining encouraging results, while the proposed modelling methodology was accepted by the experts as it was transparent, while the modelling procedure was based on causal relationships between the factors.

Fuzzy Cognitive Maps is a new promising modeling methodology, based on human knowledge. They are capable of efficiently integrating human knowledge for system modelling (Kosko 1992), dealing with ambiguity and uncertainty, and learning how to adapt to the unknown or changing environments, thus, achieving a better performance.

The proposed MDSS makes use of clinical data, doctor's opinions and suggestions and promotes the collaboration between the physicians. Its target is to model human knowledge and experience to deal with the diagnosis of CAD.

## 2 Materials and Methods

2.1 The Basic Aspects of Fuzzy Logic and Fuzzy Cognitive Maps

The term Fuzzy Cognitive Map (FCM) was introduced by Kosko (1986) to portray a cognitive map with two noteworthy characteristics: (a) the causal connections between the nodes are fuzzy and uncertain, and (b) the framework behaves in a dynamical way, where the impact of alteration in one concept influences other concepts, which in turn can influence the output. The FCM structure is comparable to that of an Artificial Neural Network, where concepts are spoken to by neurons, while the causal connections are represented by weighted joins interfacing the neurons.

A Fuzzy Cognitive Map is a representation of a belief system in a given domain. It comprises of concepts (C) representing key drivers of the system, joined by directional edges of connections (w) representing causal relationships between concepts. Each connection is assigned a weight $w_{ij}$, which quantizes the strength of the causal relationship between the concepts Ci and Cj (Dickerson and Kosko 1994).

A positive weight demonstrates an excitatory relationship, i.e., as $C_i$ increments, $C_j$ also increments, whereas a negative weight shows an inhibitory relationship, i.e., as $C_i$ increments, $C_j$ diminishes. In its graphical frame, a FCM provides a visualization of knowledge, as a collection of "circles" and "arrows", which are relatively simple to imagine and control. Key to the apparatus is its potential to permit connections among its nodes, empowering its application that alters over time. It is especially suited for use in soft-knowledge domains, where knowledge is presented qualitatively (Peng 2017).

Figure 1 shows a FCM consisting of several concepts; some of them are input concepts, whereas the rest are decision (output) concepts. Their fuzzy interactions are also depicted. The main objective of building a FCM around a problem is to be able to predict the outcome by letting the relevant nodes interact.

The concepts $C_1$, $C_2$, … $C_n$, represent the drivers and constraints that are considered of importance to the issue under consideration. The link strength between two nodes as denoted by $w_{ij}$, takes values within [-1,1]. If the value of this link takes on discrete values in the set {-1, 0, 1}, it is called a simple or crisp FCM. The concept values of nodes $C_1$, $C_2$… , $C_n$ together represent the state vector V.

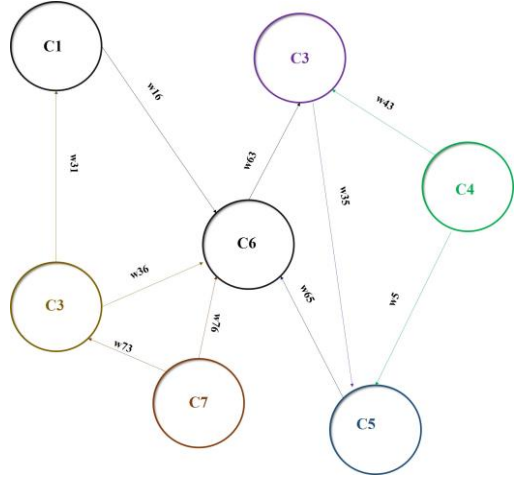

**Fig.1** Fuzzy Cognitive Map

The function describing the system involves letting the vector V evolve. The state vector V is passed repeatedly through the FCM connection matrix W. This involves multiplying V by W, and then transforming the result as follows:

$$\boldsymbol{V} = f(\boldsymbol{V} + \boldsymbol{V} \cdot \boldsymbol{W}) \quad (1)$$

$$V_i(t+1) = f\left[V_i(t) + \sum_{\substack{j=1 \\ j \neq i}}^{N} V_j(t) \cdot W_{ji}\right] \quad (2)$$

where $V_i(t)$ is the value of concept $C_i$ at step t, $V_j(t)$ is the value of concept $C_j$ at step t, $W_{ji}$ is the weight of the interconnection from concept $C_j$ to concept $C_i$ and $f$ is the threshold function squashing the result of the multiplication in the interval [0, 1], (Papageorgiou et al. 2006). We use the function:

$$f_{(x)} = \tan h(x) \quad (3)$$

In the above function, x is the value of $V_i$ at the equilibrium point.

2. 2 Development of the Decision Support System using Fuzzy Cognitive Maps

The creation and advancement of the FCM is based on on the experts' gathered information and encounter. We select the depiction of the framework to be made not by a unique expertise, but by a bunch of specialists, to portray the framework as more dispassionately as possible. The specialists portray the system's behaviour as a set of ideas, in each of which they will allot its concepts. Besides, they portray the existing relations among these ideas as cause and impact relations among the concepts.

The experts' procedure for a developing Fuzzy Cognitive Map is the following:

1. The specialists discuss and define the number and type of the concepts, which will be able to describe the main features of the system, and which will constitute the Fuzzy Cognitive Map.

2. Each specialist, in separate, defines the interaction of each system concept, according to his opinion.

3. Each specialist, in separate again, decides about the type of interaction among the concepts, namely if there will be a positive interaction $W_{ij}>0$, a negative interaction $W_{ij}<0$ or no interaction of the concept $C_i$ to the concept $C_j$.

4. Then, the connectivity degree between two concepts is defined, namely the exact value of the weight $W_{ij}$.

Each interconnection partners the relationship between the two concepts and decides the review of causality between them. The causal inter-relationships among concepts are ordinarily defined utilizing the variable impact, which is deciphered as a variable taking values within the universe $U = [-1,1]$. Utilizing ten etymological factors, the specialists can portray in detail the impact of one concept on another and can perceive between distinctive degrees of impact. At that point, the etymological factors proposed by the specialists for each interconnection are aggregated utilizing the SUM method and an etymological weight is delivered, which is defuzzied with the Centre of Gravity method. Hence, a numerical weight for $W_{ij}$ is produced. Utilizing this strategy, all the weights of the FCM are inferred.

2.3 Coronary Artery Disease

According to the World Health Organization, 50% of deaths in the European Union are caused by cardiovascular diseases, while 80% of premature heart diseases and strokes can be prevented (Mendis et al. 2015). The American Heart Association (AHA), created a new set of central Strategic Impact Goals, in 2011, to drive organizational priorities for the current decade: "By 2020, to improve the cardiovascular health of all Americans by 20%, while reducing deaths from CVDs and stroke by 20%" (WHO 2018). Cardiovascular diseases, including strokes, affect all ages, men and women, and all social groups. 1 out of 10 men, aged 50-59, has a "silent" coronary disease and is at risk of having a heart attack without any warning (Montalescot 2013).

Coronary Artery Disease is caused when the atherosclerotic plaques load, namely fill, in the lumen of the blood vessels of the heart, which are named coronary arteries, and they obstruct the blood flow to the heart. This results to a decreased provision of oxygen and nutritional substances to the cardiac tissues (Montalescot 2013). In general, the stenosis of >70% of the vessel's diameter is considered abnormal (Willerson et al. 2007).

A unique feature of the disease that must be taken into consideration is that it typically has only one symptom, that of pain. This cannot supply the experts with adequate information as to whether the patient suffers or not, because it is a symptom that might be random (American Heart Association 2017).

Factors that contribute to an increased risk of suffering from the particular disease have to be taken into consideration in order for a diagnosis to be accurate. Physicians suggest that four groups of factors lead to the decision. Those are predisposing factors (e.g. gender, age, family history), intercurrent diseases (e.g. diabetes), diagnostic tests (e.g. stress echo) and of course, the type of pain (e.g. typical angina, atypical angina). Those will be the nodes of our model as the doctors will suggest. For each person, even the most experienced and capable, the process of evaluating many factors is unfeasible. Thus, many times, the results of the diagnostic tests are overestimated.

## 2.4 The proposed model

### 2.4.1 The concepts of the FCM

One main difference of our proposed system from other techniques, is that it trusts the doctor's opinion regarding the results and the significance of the various diagnostic tests. It shall not try to diagnose the CAD by interpreting the results of the tests (i.e., making use of measurements derived from the tests). Interpreting the result of a diagnostic test is performed by the experts, who follow specific guidelines to reach to a medical report.

Three physicians-experts were pooled to define the number and the type of parameters-factors affecting Coronary Artery Disease. Those factors are shown in Table 1.

The factor of typical angina will be excluded from the proposed MDSS, due to the fact that it requires no further examination, when present. Factors such as gender, age and diagnostic tests require different handling when taking absolute values. Therefore, several attributes (e.g., young age, normal diagnostic tests) are split, as shown in Table 1.

The output (A31) takes the potential values "Zero Probability", "Small Probability", "Medium Probability", "Relatively Large Probability", "Large Probability", and "Very Large Probability". However, in this work, the output will also take (%) values, that show a probability for the (candidate) patient to suffer from the disease.

**Table 1.** The concepts of the FCM

|       | Attributes                      | Values                                  |
|-------|---------------------------------|-----------------------------------------|
| (N/A) | typical angina pectoris         | yes, no                                 |
| A1    | atypical angina pectoris        | yes, no                                 |
| A2    | atypical thoracic pain          | yes, no                                 |
| A3    | dyspnea on exertion             | yes, no                                 |
| A4    | Asymptomatic                    | yes, no                                 |
| A5    | gender – male                   | yes, no                                 |
| A6    | gender – female                 | yes, no                                 |
| A7    | age <40                         | yes, no                                 |
| A8    | age [40-50]                     | yes, no                                 |
| A9    | age [50-60]                     | yes, no                                 |
| A10   | age >60                         | yes, no                                 |
| A11   | known cad                       | yes, no                                 |
| A12   | previous stroke                 | yes, no                                 |
| A13   | peripheral arterial disease     | yes, no                                 |
| A14   | Smoking                         | yes, occasionally, no                   |
| A15   | arterial hypertension           | yes, no                                 |
| A16   | Dyslipidemia                    | yes, no                                 |
| A17   | Obesity                         | yes, relatively, no                     |
| A18   | family history                  | yes, no                                 |
| A19   | Diabetes                        | yes, no                                 |
| A20   | chronic kidney failure          | yes, no                                 |
| A21   | electrocardiogram normal        | yes, no                                 |
| A22   | electrocardiogram abnormal      | yes, no                                 |
| A23   | echocardiogram normal - doubtful| yes, no                                 |
| A24   | echocardiogram abnormal         | little, abnormal, definitely abnormal   |
| A25   | treadmill exercise test normal  | yes, no                                 |
| A26   | treadmill exercise test abnormal| abnormal, definitely abnormal           |
| A27   | dynamic echocardiogram normal   | yes, no                                 |
| A28   | dynamic echocardiogram abnormal | doubtful, abnormal, definitely abnormal |
| A29   | scintigraphy normal - doubtful  | yes, no                                 |
| A30   | scintigraphy abnormal           | little, abnormal, definitely abnormal   |
| A31 | prediction of infection | zero, small, medium, rel. large, large, very large |

*2.4.2 Development of the weight Table*

Each variable affects the outcome with a corresponding weight. However, there are also internal relations among these variables. Note that the ability for each expert to define its own internal connections has not been predicted by this specific system. The rules presented above are rules proposed by all three doctors. These inter-relations are demonstrated in Table 2. The possible variables that an interconnection weight can take are described as follows:

- VW (very weak): the relation between the concept Ai and Aj is very weak
- W (weak): the relation between the concept Ai and Aj is weak
- M (medium): the relation between the concept Ai and Aj is medium
- S (strong): the relation between the two concepts is strong
- VS (very strong): the relation between the two concepts is very strong
- The highest Possible (THP)

**Table 2**. Forever standing relationships between concepts

| Rule | Attributes Affected - Affection |
|---|---|
| **If woman** | Age [40 50] – sharply reduced impact, Age [50 60] – medium reduced impact, Diagnostic Tests – average increase at the weights of normal and medium decrease at the weights of abnormal ones. |
| **If Definitely Abnormal Scintigraphy** | The weight of Scintigraphy is significantly increased |
| **If ECG is Normal and Scintigraphy is Normal** | The weights of both tests are relatively increased |
| **If previous Stroke** | Negate the Gender Discrimination and de-activate the attribute from the system |
| **If Known Cad** | Negate the family history affection |

Accordingly, the affection of the rules that were illustrated on the Table 2 will be translated into equal weight – rules. For example, the rule "The weight of Scintigraphy is greatly increased" will be translated into a change of the weight from M (Medium) to (VS) Very Strong. If the weight is already Very Strong and the doctor suggests a further increase, then a mathematic procedure to handle with this will be implied. In Table 3, the doctors' opinions of the weights that associate each attribute with the diagnosis (A31) are presented. We present here an example of the linguistic values the doctors proposed. The reader should note that the weights referred in Table 3 express the relation between the attributes and the output (A31).

**Table 3.** Example of weight definition by doctors

| Attribute | Doctor 1 | Doctor 2 | Doctor 3 |
|---|---|---|---|
| A28 | -M | -M | -M |
| A9 | -VW | -W | 0 |
| A29 | (-)THP | (-)VS | (-)S |
| A30 | THP | THP | THP |

The linguistic values of the weights are transformed into one numerical value for each expert and then into one value for each interconnection, as presented in Table 4.

**Table 4.** Defuzzied weight values

|     | Attributes | Weights |     | Attributes | Weights |
| --- | --- | --- | --- | --- | --- |
| A1  | Atypical Angina Pectoris | 0.35 | A17 | Obesity | 0.2 |
| A2  | Atypical Thoracic Pain | 0.2 | A18 | Family history | 0.1 |
| A3  | Dyspnea on exertion | 0.25 | A19 | Diabetes | 0.4 |
| A4  | Asymptomatic | -0.35 | A20 | Chronic Kidney Failure | 0.15 |
| A5  | Gender – Male | 0.3 | A21 | Electrocardiogram Normal | -0.4 |
| A6  | Gender - Female | -0.5 | A22 | Electrocardiogram Abnormal | 0.35 |
| A7  | Age <40 | -0.75 | A23 | Echocardiogram Normal - Doubtful | -0.35 |
| A8  | Age [40-50] | -0.25 | A24 | Echocardiogram Abnormal | 0.42 |
| A9  | Age [50-60] | 0.1 | A25 | Treadmill Exercise Test Normal | -0.75 |
| A10 | Age >60 | 0.4 | A26 | Treadmill Exercise Test Abnormal | 0.6 |
| A11 | Known CAD | 0.35 | A27 | Dynamic Echocardiogram Normal | -0.44 |
| A12 | Previous Stroke | 0.1 | A28 | Dynamic Echocardiogram Abnormal | 0.625 |
| A13 | Peripheral Arterial Disease | 0.1 | A29 | Scintigraphy Normal - Doubtful | -0.85 |
| A14 | Smoking | 0.1 | A30 | Scintigraphy Abnormal | 0.7 |
| A15 | Arterial Hypertension | 0.1 | A31 | Prediction of infection |     |
| A16 | Dyslipidemia | 0.15 |     |     |     |

The developed network in a block diagram format is shown in Figure 2.

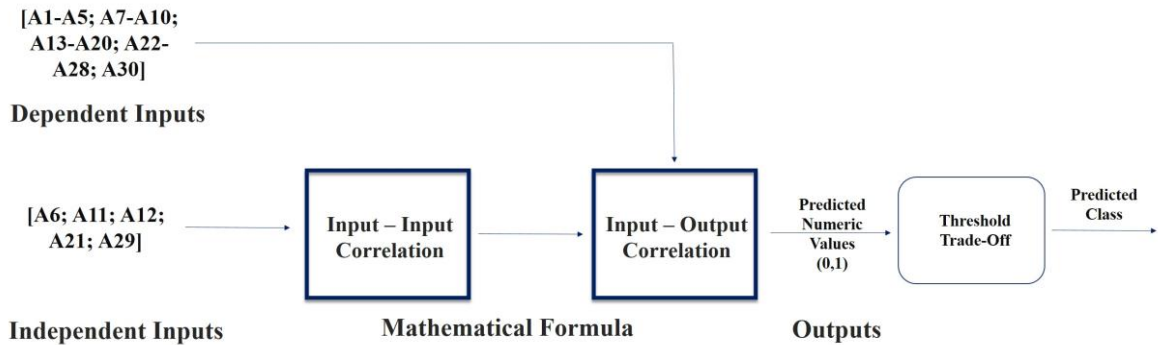

**Fig. 2.** Block diagram of the developed FCM

*2.4.3 Mathematical Procedure*

Using equation (2) and the table of weights, we can reach a decision in a mathematical expression. Equation 2, when fed with a specific vector **V,** which represents the initial state of the system, and when the weight vector is applied, is reaching to a steady-state. Each element of the vector describes the final state of each attribute. The last element is the summary of each attribute's affection on the prediction. For a new case inserted as an input to the above system, the final state of Vector **V** is a unique number in space [-3, 6] before the application of the normalization function $f$ (eq. 3). This number reflects the probability of this case to be suffering from the disease.

We can use the normalization function to transfer this number in space [0,1]. In this work, we used the tanh function to normalize the values is space [-1,1].

*2.4.5 Fuzzification of the Output*

The potential values "Zero Probability", "Small Probability", "Medium Probability", "Relatively Large Probability", "Large Probability", and "Very Large Probability" must be derived from a certain space. In this work we chose those spaces to be:

- If $V_{31} \ni [-1,0]$ then "Zero Probability"
- If $V_{31} \ni [0,0.25]$ then "Small Probability"
- If $V_{31} \ni [0.25,0.5]$ then "Medium Probability"
- If $V_{31} \ni [0.5,0.75]$ then "Large Probability"
- If $V_{31} \ni [0.75,1]$ then "Very large Probability"

2.5 Activation function

As demonstrated above, the proposed MDSS and the fuzzification technique is designed to inform the doctor of a probability in a linguistic fuzzy way (e.g. "High Probability"). In order to evaluate the system using real labelled data, we have to force the system to classify every instance between "diseased" or "healthy". One second option was to ignore instances corresponding to the fuzzy outputs "Medium Probability" thereby considering this zone as a grey one.

However, as all of the classifiers we employ for comparisons are following the same procedure (no grey zones), the proposed MDSS will classify every instance. We extensively evaluate the appropriate threshold to achieve the best trade-off between the desired Sensitivity and Specificity, which are common measurements related to medical tests.

2.6 Dataset of the study

The database utilized in this work consisted of 303 patient cases, recorded at the Department of Nuclear Medicine of Patras, in Greece. The majority of instances were recorded at the last 8 years. All the patient cases had been pointed to Surgical Coronary Angiography to confirm the affection of CAD. In order for a doctor to characterize an instance as healthy or disease, the stenosis of the coronary artery is the only criterion, which is obtained by the above mentioned invasive diagnostic test. Stenosis equal or above 70% were labelled as diseased, whereas above 70% were labelled as healthy. The data we used for the experiments was anonymous.

The dataset contains 116 healthy cases and 187 diseased cases. The dataset consists of 266 male instances and 37 female ones. The attributes of our dataset are corresponding to the factors influencing the diagnosis of CAD, as they were described above. The factors include: the type of symptom (atypical angina, atypical thoracic pain, dyspnea on exertion, asymptomatic), age, gender, predisposing factors and recurrent diseases (Known Cad, Family History, Obesity, Diabetes, Arterial Hypertension, Dyslipidemia, Kidney failure, Peripheral Arterial Disease) and the results of the diagnostic tests (ECG, Dobutamine Stress Test, Stress Echo, Treadmill Exercise Test and Scintigraphy).

The medical reports regarding the diagnostic test were translated in numeric values with the appropriate staging. The medical staff supervised this process. The attributes regarding the patient's history and condition (i.e. smoking) were also in need of some processing in order to turn the linguistic values (i.e. "smoker") into zeros and ones.

2.7 Evaluation Criteria

We extensively evaluate our system, therefore, more criteria besides accuracy will be employed. The evaluation criteria of the system will be the following: (a) Accuracy based on the whole dataset, (b) True Positives, (c) False Positives, (d) False Negatives, (e) True Negatives, (f) sensitivity, (g) specificity, (h) Negative Predict Value (NPV) and (i) Positive Predict Value (PPV). We consider the proposed MDSS seriously problematic, if the number of false negatives is exceeding an acceptable medical threshold.

# 3 Results

3.1. Classification accuracy

The results are illustrated in Table 5. The confusion matrix corresponds to sensitivity of 83.96%, specificity of 68.97%, PPV of 81.34% and NPV of 72.73%. The overall accuracy is 78.21%.

**Table 5.** Confusion Matrix

|  | Actual CAD | Actual Normal | Total |
| --- | --- | --- | --- |
| **Predicted CAD** | 157 | 36 | 193 |
| **Predicted Normal** | 30 | 80 | 110 |
| **Total** | 187 | 116 | 303 |

Considering the fact that the proposed MDSS was not trained in any way on the dataset, the results are more than encouraging. The results show that the system provides good sensitivity and NPV as well as PPV. However, its specificity is average. That is due to the high number of False Positives. This derives from the database itself, as it does not contain an equal amount of Diseased and Healthy patients. The reader should note that the corresponding accuracy of the medical reports (before the Coronary Angiography test) is approximately the same (~77%), which demonstrated that the proposed system competes with the experts' predictions.

3.2 Comparisons

The dataset was also utilized to train common Machine Learning algorithms. Two methods of making use of the dataset for training and testing were used (5-fold cross-validation and 70%-30% train and test data). For the Machine Learning algorithms, the Waikato Environment for Knowledge Analysis, developed at the University of Waikato in New Zealand, was employed. We briefly describe the parameters of the algorithms applied to the dataset.

The Neural Network contains 1024 hidden layers and is trained for 120 epochs, with a batch size of 32 and an initial learning rate of 0.01. Other Neural Networks with alterations on the pre-mentioned parameters were tested and excluded due to inefficiency. Hence, the Neural Network we pick is the one with the better accuracy over all. The Support Vector Machine (SMO) has the following parameters: Complexity Parameter (c) is set at 1.0, the value of epsilon is set at 1E-12, the batch size is 32 and the calibrator method is tuned to Logistic. AdaBoostM1 meta classifier is trained for 50 iterations, with a batch size of 32 and the decision classifier Decision Stump is selected. Chirp is trained with a batch size of 32. Random Forest was trained for 100 iterations with a bag size of 20%, and batch size 32. Spaarc was trained with a batch size of 32.

In Table 6 we present the results of state-of-the-art classifiers, trained and tested on the dataset of the study. Several other classifiers were trained, but their results fell far below 60% at either 5 fold cross-validation or data split, and were not examined extensively. The results demonstrate that the proposed model, archives at least 2% better accuracy compared to the best accuracy obtained from the experiments (that is with Chirp Classifier).

**Table 6.** Classifier Results

| Classifier | Accuracy (5-fold – cross validation) | Accuracy (70-30) |
|---|---|---|
| Coarse Tree | 63.4 | 65.5 |
| Linear Discriminant | 70.3 | 70.9 |
| Logistic Regression | 70.6 | 68.9 |
| Linear SVM | 70.0 | 72.2 |
| Cubic SVM | 68.0 | 69.5 |
| Medium Gaussian | 73.3 | 70.9 |
| Coarse Gaussian | 62.7 | 61.6 |
| Medium KNN | 67.7 | 68.2 |
| Cubic KNN | 66.3 | 68.9 |
| Ensemble Bagged Trees | 68.3 | 68.2 |
| Ensemble Subspace Discriminant | 71 | 71.5 |
| Neural Network | 72.6 | 71.42 |
| Support Vector Machine (SMO) | 72.93 | 71.42 |
| AdaBoostM1 | 74.58 | **75.82** |
| Chirp | **76.89** | 72.52 |
| Spaarc | 72.93 | 74.72 |
| Random Forest | 74.58 | 71.42 |

## 4 Discussion

In this paper, a very challenging health problem of the medical professionis considered and studied using engineering and control theories.

The proposed decision support system achieve its aim under the conducted experiments. In the total database, a 78.2% accuracy is competing with the accuracy of the medical staff. Besides, the classification accuracy is exceeding the diagnostic accuracy of the medical tests incorporated into the dataset. Specifically, the most reliable diagnostic test (Scintigraphy) obtained an accuracy of ~70% in the particular dataset, while the sensitivity and specificity were analogously lower. Besides, the proposed algorithm, outperformed several state-of-the-art Machine Learning algorithms, which were proven to be inefficient for the particular dataset's size and complexity.

There are some drawbacks the proposed system has to overcome in future research. The main issue is the fact that all the concepts are directly affecting the output concepts, regardless of their different nature. This is not scientifically accepted. The second issue is the fact that the output seldom was triggered by every possible concept. That is due to missing data on certain diagnostic tests. For this reason, applying the same normalization function to every new instance is not actually placing the values to the desired space. Recent advances in FCM proposed by Mpelogianni et al. (2018), and by Groumpos (2018) may be utilized to overcome those issues.

In conclusion, the contribution of this study lies in demonstrating the effectiveness of Fuzzy Cognitive control in medical diagnostic tasks, and, especially is uncertain conditions with absence of large-scale data.